# Pedestrian Motion State Estimation From 2D Pose

Fei Li, Shiwei Fan, Pengzhen Chen, and Xiangxu Li

*Abstract*—Traffic violation and the flexible and changeable nature of pedestrians make it more difficult to predict pedestrian behavior or intention, which might be a potential safety hazard on the road. Pedestrian motion state (such as walking and standing) directly affects or reflects its intention. In combination with pedestrian motion state and other influencing factors, pedestrian intention can be predicted to avoid unnecessary accidents. In this paper, pedestrian is treated as non-rigid object, which can be represented by a set of two-dimensional key points, and the movement of key point relative to the torso is introduced as micro motion. Static and dynamic micro motion features, such as position, angle and distance, and their differential calculations in time domain, are used to describe its motion pattern. Gated recurrent neural network based seq2seq model is used to learn the dependence of motion state transition on previous information, finally the pedestrian motion state is estimated via a softmax classifier. The proposed method only needs the previous hidden state of GRU and current feature to evaluate the probability of current motion state, and it is computation efficient to deploy on vehicles. This paper verifies the proposed algorithm on the JAAD public dataset, and the accuracy is improved by 11.6% compared with the existing method.

## I. Introduction

According to several public reports, thousands of pedestrians die each year from traffic accidents. In order to improve people's travel safety, intelligent vehicles are getting more and more attention. An important case for self-driving vehicles is that they can reduce the number of traffic accidents.

Vulnerable Road Users (VRUs), such as pedestrians, are the most challenging road users to deal with. Pedestrian can rapidly switch between various motion modes, such as walking or standing. Pedestrian detection and tracking usually depends on visual perception [1-3]. The result of visual perception is mainly bounding box, its corresponding box center, and class. While position, velocity, and other information of pedestrian are obtained through fusion with perception results of other sensors, such as lidar. Pedestrian prediction module receives information of pedestrian perception module and predicts intention or trajectory in the future. Separate models are often learned for each of the identified motion state [4]. The various information needs to be integrated into a predictive model to assess which state the pedestrian will switch to, and when the pedestrian will do so. Pedestrian detection, tracking and prediction are some of the most challenging tasks in intelligent vehicles field due to the following characteristics of pedestrian.

- High mobility：The high mobility of pedestrians is mainly reflected in the transient state of pedestrian movement, such as the instantaneous switching of walking and standing, and the variability of pedestrian movement direction; this characteristic not only affects the accuracy of pedestrian perception, but also deteriorates the multi-modal problem of pedestrian predictions to a certain extent, which makes pedestrian predictions more difficult.

- Non-rigid: Pedestrian is non-rigid, so bounding box is time-varying due to the micro-motion of legs and arms relative to the torso, such as arms swinging. The time-varying bounding box causes random shift in location of pedestrian.

- Small and low speed: Although there are cameras, lidars, and millimeter-wave radars in autonomous driving, the perception of small and low-speed pedestrian is still a very challenging research under a world coordinate system (not image coordinate system). The signal-to-noise ratio in pedestrian perception results is much worser than that of vehicles, and so tracking stability does.

Due to such characteristics of pedestrians, position and velocity information estimated by the perception module are seriously corrupted by noise, and the performance of pedestrian prediction is limited by this corrupted information. Existing end-to-end pedestrian prediction algorithms [5-11] usually require long-term historical trajectories. As discussed above the pedestrian tracking in the autonomous driving scene is not robust enough to obtain long-term tracking information, so pedestrian prediction based on short-term historical information even current information is the more feasible method currently [12-13]. Pedestrian motion state can effectively improve the stability of pedestrian tracking [14-15]. In addition, a pedestrian motion state directly affects its behavior or intention. Combining pedestrian motion state and other information can improve the performance of predicting pedestrian intention and trajectory, especially with short-term historical information [12-13]. Therefore, the motion state of pedestrian is of great significance for pedestrian tracking and prediction. In fact, in [8] it is concluded that a lack of information about the pedestrian's posture and body movement results in a delayed detection of the pedestrians changing their crossing intention.

There are several pedestrian motion state estimation methods summarized as follows:

- Motion state estimation based on the current and previous tracking information: This method obtain velocity from previous observations and estimates the motion state. While previous information does not

Fei Li, Shiwei Fan, Pengzhen Chen, and Xiangxu Li are with Noah's Ark Lab, Huawei. (e-mail: lifei120@huawei.com, fanshiwei@huawei.com, chenpengzhen@huawei.com, lixiangxu@ huawei. com).

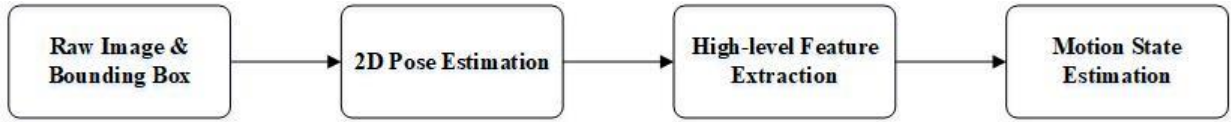

Figure 1. Overall framework of proposed method

have strong constraints on pedestrian motion due to the high mobility of pedestrian, so this method is not reliable in some circumstance. This method has low computational complexity but poor performance, and is affected by noise in pedestrian perception results.

- Motion state estimation based on visual images: This type of method usually uses a general CNN based backbone network to extract features, and then classify pedestrian motion state via fully connected layer. Due to the use of original visual features, the performance of such methods is easily affected by environmental factors and occlusions. In [7] they use a pre-trained AlexNet [16] on ImageNet [17], and transform the last three fully connected layers (fc6-fc8) to convolution layers with a global max-pooling layer at the output to generate the final class scores. This method achieves 83.45% average precision of walking state with one single frame image. This method does not take temporal information into account.

- Motion state estimation based on key point or pose information [18-19]: In [18], 3D key points are extracted from a stereo pair, and B-GPDMs (balanced Gaussian process dynamical models) are utilized for reducing the 3-D time-related information extracted from key points or joints placed along pedestrian bodies into low-dimensional spaces，and multiple models are learned for each type of pedestrian activity, and selects the most appropriate one to estimate pedestrian states.

Seq2seq model is utilized here and trained end to end for motion state estimation, such as walking and standing. Instead of low-level semantic feature representation used by image based method, we utilize 2D pose information to enhance feature representation explicitly. By extracting high-level features based on human pose, pedestrian is treated as non-rigid target, static and dynamic micro motion features are used to represent motion pattern. Micro motion means the motion of keypoints or joints relative to the torso. Micro motion helps to obtain accurate motion state when pedestrian speed is low or motion state changes. Seq2seq model is used to learn the dependency across feature sequence or the transition pattern between different motion states, final class probabilities is easy to obtain by a simple classifier. Note that, our method only relies on a monocular stream of images, these other methods [18-19] cannot be applied due to the lack of stereo information and vehicle data for ego-motion compensation.

The main contributions of this paper are as follows: 1) seq2seq model: Based on the characteristics of high mobility, the seq2seq model is used to learn the dependence of pedestrian motion state transition on previous information. 2) Static and dynamic micro-motion features: In the target tracking model, pedestrians are usually regarded as rigid objects or point targets. The micro-motion features is used to describe the minor movement of pedestrian to distinct different motion states. These features are effective and sensitive to the movement and state transition of pedestrian. 3) We perform experiments on JAAD dataset, which is acquired in naturalistic conditions and annotated, to evaluate the performance of the proposed algorithm against baseline.

This paper is organized as follows. Section II presents the proposed method, including the methods of high-level feature extraction and network structure. Section III shows experimental results on JAAD datasets. Conclusions are given in Section IV.

## II. PROPOSED METHOD

For our part of the research, we would like to focus only on the reasoning part of the problem, which means it is assumed that the pedestrian detection has already been addressed by a generic pedestrian detector. The proposed method receives a bounding box of a pedestrian from an image or a single video frame as input, and outputs the probabilities for each motion state.

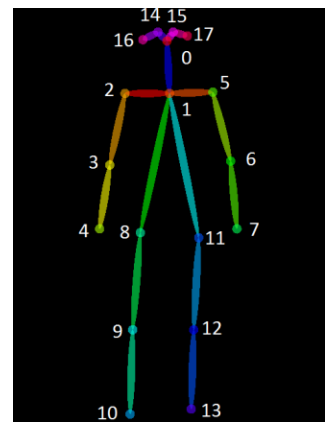

Figure 2. Illustration of keypoint in COCO format

The method proposed in this paper is composed of three main steps, as shown in Figure 1. First, an existing pose estimation software [20-21] is utilized to obtain 2D pose information given bounding box of pedestrian, and then high-level features, static micro motion features and dynamic micro motion features are extracted from pedestrian joint locations. Finally, seq2seq based network is trained to model the motion state transition, and learns the dependency across time series, probabilities of different motion state are calculated by a softmax classifier.

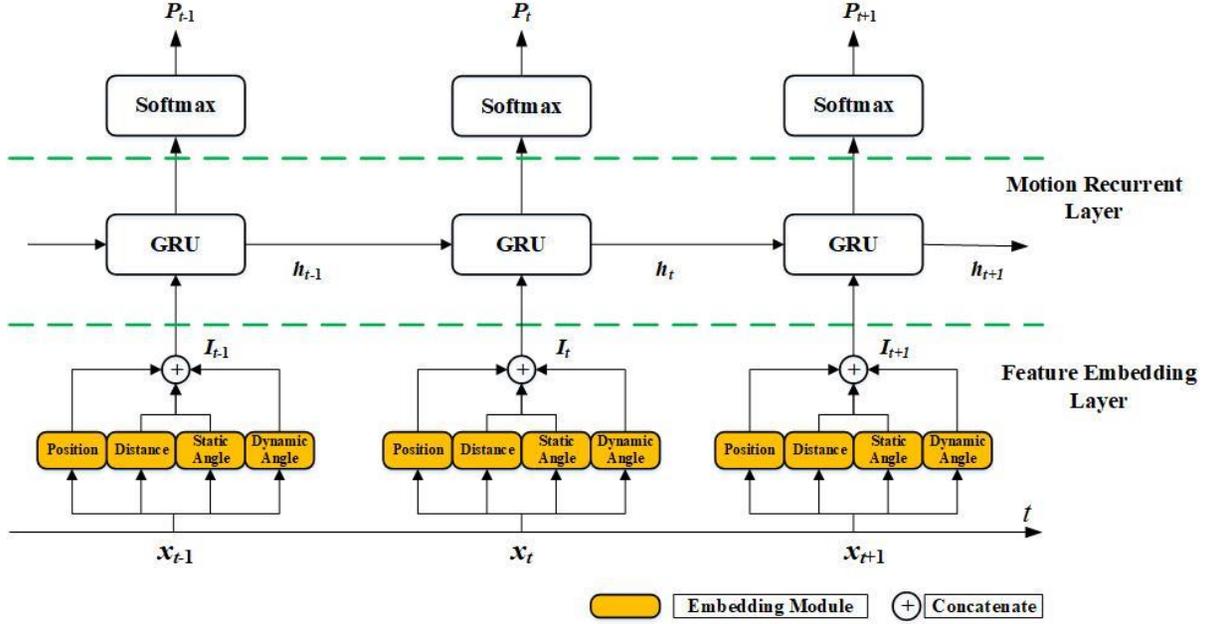

Figure 3. Architecture of our proposed model for pedestrian state estimation. At each time step, different type of micro motion feature is encoded separately and concatenated to generate an internal representation $I_t$. Afterwards, the motion recurrent layer works on the internal representation and last hidden state to yield motion state representation. Finally, softmax classifier calculate probability of each motion state. $x_t$ is the input feature vector at time step $t$, $P_t$ is the probabilities of different motion states at time step $t$. Yellow rectangle represents an embedding module.

### A. 2D Pose Estimation

For multi-person pose estimation, there are two categories of 2D pose estimation: top-down fashion and bottom-up fashion. In a top-down fasion, the human pose estimation algorithm is applied to estimate the locations of key points, such as shoulders, elbows, wrists, hips, knees and ankles, given each pedestrian's bounding box obtained by a generic pedestrian detector [20-21]. In contrast, in a bottom-up fashion, human key point candidates are first detected and then associated to individual human [22-23].

Here, an existing pose estimator, AlphaPose [20-21], is adopted to obtain noisy keypoints in COCO format, as shown in Figure 2. With the observations of keypoints, pedestrian motion state can be estimated as described below.

### B. High-Level Feature Extraction

This subsection describes the high-level features which consist of static and dynamic micro-motion features, extracted from locations of 2D keypoints. Here, micro-motion refers to the movement of the pedestrian's joints relative to the torso, which is different from the traditional macro-movement of a rigid target or a point target.

1) Static micro motion feature

Micro motion of pedestrian's joints induces changes not only in the position of key points, direction of joints, but also in the angle between different joints. Static micro-motion features include the normalized position of key points, the normalized distance between different key points, and the direction of different joints.

**Normalized position:** This kind of feature consists of the coordinates of arms and legs. First, the coordinates are translated to a coordinate system with the neck point (key point 1 shown in Figure 1) as the origin. Then position is normalized with the height of the bounding box determined by the key points.

**Normalized distance:** Normalized distance features mainly include Euclidean distances between different key points and its components on two coordinate axes. The distance features of legs include the distance between left and right ankles (key point 10 and key pint 13), and between left and right knees (key point 9 and key pint 12). The distance features of arms include the distance between left and right hands (key point 4 and key pint 7), and between left and right elbows (key point 3 and key pint 6). Hip width (key point 8 and key pint 11) is used to normalize the Euclidean distance features, and height is used to normalize the components of the Euclidean distance on each coordinate axis.

**Angle features:** Angle features include the angles between each joint of limbs and the horizontal axis, and the angle between each pair limb joints, such as the angle between left forearm and the horizontal axis, the angle between left forearm and right forearm. It also includes the angle of each two key points in the limbs with respect to the horizontal axis, such as direction from right hand to left elbow relative to horizontal axis.

2) Dynamic micro motion feature

Dynamic micro motion features are differential calculations of static micro motion features in time domain, mainly consisting of dynamic Euclidean distance features and dynamic angle features.

## C. Motion State Estimation

We address motion state estimation as the search for a function that maps an input sequence to an output sequence. So seq2seq model is adopted to learn the mapping function and state transition pattern. Architecture of our proposed model for pedestrian state estimation is shown in Figure 3. Static and dynamic micro-motion features $x_t$ mentioned above are divided into four types of features: position features, distance features (static and dynamic distance features), static and dynamic angle features. Considering the differences in dimensions and physical meaning between different features, these features are encoded separately with different embedding layers which are fully connected layers with 16 hidden units. Then, the outputs of feature embedding layer are concatenated together, referred as internal representation $I_t$, and sent to motion recurrent layer. At each time step, the current motion state is estimated based on the combination of inputs at current time and previous information, not the whole sequence. So when our model infers online, inputs of our model include features at current time and previous state. For the modeling of our sequence data, we use gated recurrent units (GRUs) [24] which are simpler compared to LSTMs and, in our case, achieve similar performance. GRU uses a gating mechanism to control input, memory and other information to make predictions at the current time step. The expression is given below.

$$r_t = \sigma(W_{rx}f(x_t) + W_{rh}h_{t-1}) \quad (1)$$

$$z_t = \sigma(W_{zx}f(x_t) + W_{zh}h_{t-1}) \quad (2)$$

$$\tilde{h}_t = tanh(W_{xh}f(x_t) + W_{hh}(r_t \odot h_{t-1})) \quad (3)$$

$$h_t = (1 - z_t) \odot h_{t-1} + z_t \odot \tilde{h}_t \quad (4)$$

where $x_t$ is the input feature vector at time step $t$, $f(\cdot)$ is the embedding function, and $f(x_t)$ is the internal representation $I_t$; $\sigma(\cdot)$ is the sigmoid function, $r_t$ and $z_t$ are reset and update gates, and matrices $W$ are weights between two units.

The final scores $P_t$ of different motion states are generated by the softmax classifier. The motion recurrent layer learns the state-dependent relationship at different times and the motion state transition function by considering previous state information and current time input.

## D. Implementation

In our architecture, position embedding, distance embedding, static and dynamic angle embedding are fully connected layers, each with 16 hidden units, and we use GRU with 64 hidden units. All the activation function is set to tanh. Batch normalization is added before the nonlinear transform in feature embedding layer.

**Training:** The model is trained using ADAM [25] optimizer with an initial learning rate of 0.0002 for 80 epochs, with batch size of 32 and learning rate is exponentially decayed every 3000 updates with a coefficient of 0.9. The maximum length of sequence is set to 64, and minimum length is 30. To alleviate over fitting, L2 regularization with weight 0.0005 is added in loss function, a dropout layer is also added after feature concatenation with a rate of 0.5.

## III. EXPERIMENTS AND RESULTS

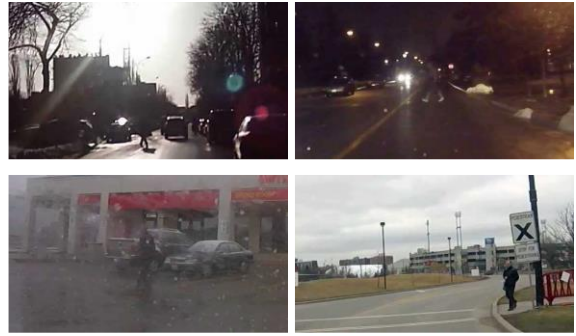

Figure 4. Samples of visibility changes in our dataset. These examples show how weather/lighting conditions can make the estimation of pedestrians motion state challenging.

### A. Datasets

JAAD [7] is a dataset for studying joint attention in the context of autonomous driving. In particular, the focus is on pedestrian and driver behaviors at the point of crossing and the factors that affect them. To this end, the JAAD dataset provides a collection of annotated short video clips representing typical urban driving scenarios under various weather conditions. Figure. 4 shows some examples of changes in visibility, and how challenging pedestrian motion state estimation can be in some cases.

TABLE I. PROPERTIES OF JAAD DATASET

| | |
|---|---|
| # of frames | 82K |
| # of annotated frames | 75K |
| # of pedestrians | 2.8K |
| # of pedestrians with behavior annot | 686 |
| # of pedestrian bboxes | 391K |
| Avg. pedestrian track length | 140 frames |

There are over 300 video clips in JAAD dataset ranging from 5 to 15 seconds in duration. Table I summarizes the properties of JAAD dataset. The data was collected in North America (60 clips) and Europe (286 clips) using three different high-resolution monocular cameras. The cameras were positioned inside the car below the rear view mirror. The frame rate of the videos is 30 fps.

JAAD dataset was divided into training, testing and validation. We use training and validation dataset for training our model and testing dataset for testing. Under this setting, training data includes 56950 frames of walking sample, 11007 frames of standing sample, and testing data includes 43725 frames of walking sample and 6669 frames of standing sample. We augment the data at training time by horizontally flipping the images and sub-sampling the over-represented class to equalize the number of standing and walking samples.

## B. Metrics

We report all the evaluation results using the following metrics: Accuracy, F1 score, precision and recall.

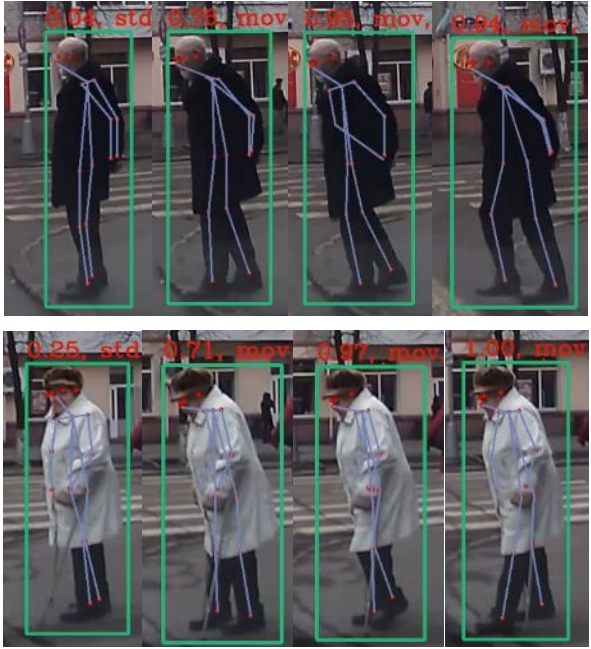

Figure 5. Examples of motion state transition, the decimal is the probability of walking, and the string is the ground truth of motion state at current time.

## C. Experimental results

We evaluate the performance of our proposed algorithm against the baseline model in [7]. In [7], AlexNet pre-trained on ImageNet is fine-tuned with cropping the bottom half of the bounding box for motion state estimation. The comparison results are summarized in Table II. Our 2D pose-based approach outperforms previously proposed approach by a large margin carried out on the JAAD dataset with 11.6% improvement in precision. From Table II, we can see that static and dynamic micro-motion features extracted from 2D pose can describe pedestrian motion precisely under various conditions, and previous hidden state helps to learn the transition relationship between different motion states.

TABLE II. COMPARISON WITH BASELINE

| Method | Performance | | | |
|---|---|---|---|---|
| | Precision | Accuracy | F1 score | Recall |
| Baseline [4] | 0.835 | None | None | None |
| Ours | 0.951 | 0.887 | 0.92 | 0.89 |

Some examples are shown in Figure 5 to illustrate the effectiveness in motion state transition case. Proposed method can estimate motion state accurately during this transition procedure, and this can prove our seq2seq model learned some motion state transition patterns. Probabilities of motion state, ground truth of motion state and crossing intention are shown in Figure 6. From Figure 6, we can see that our model is stable at estimating motion state when state varies. Furthermore, intention is closely related to motion state, and we can predict pedestrian intention in advance with other factors such as body orientation and distance to curb.

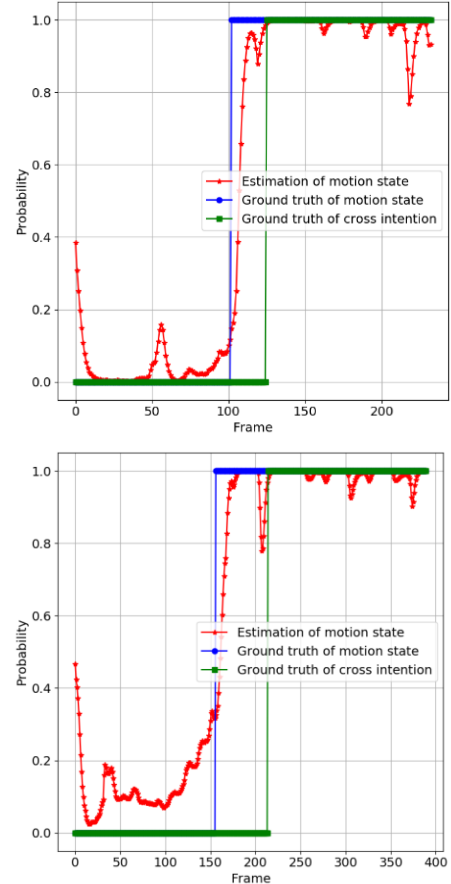

Figure 6. Motion state and intention corresponding to Figure 5, intention means crossing.

TABLE III. COMPARISON WITH BASELINE

| Ablation experiments | Precision | Accuracy | F1 score | Recall |
|---|---|---|---|---|
| without position features | 0.944 | 0.873 | 0.909 | 0.877 |
| without distance features | 0.946 | 0.87 | 0.907 | 0.871 |
| without angle features | 0.942 | 0.851 | 0.892 | 0.846 |
| without dynamic features | 0.905 | 0.868 | 0.905 | 0.874 |
| with all features | 0.951 | 0.887 | 0.92 | 0.89 |

## D. Ablation Study

To verify the effectiveness of different micro motion features designed in II-B, ablative experiments are designed. The proposed model was trained without distance feature, angle feature, position feature and dynamic feature respectively, compared with the model trained with all features. Comparison of contribution of different groups of micro motion features is demonstrated in TABLE III. This comparison shows that all groups of features extracted from 2D pose are effective and essential for the estimation of motion state.

## IV. CONCLUSION

Aiming at motion state estimation from a single frame image, a seq2seq model is designed to learn the transition relationship between different motion states with high-level micro-motion features extracted from 2D pose. Experiments conducted over JAAD dataset show that the proposed method outperforms baseline by a large margin for the task of pedestrian motion state estimation. Future work may focus on predicting intention or trajectory of pedestrians with motion state.

## REFERENCES

[1] Hasan I, Liao S, Li J, et al. Pedestrian Detection: The Elephant In The Room[J]. arXiv preprint arXiv:2003.08799, 2020.
[2] Wang W. Adapted Center and Scale Prediction: More Stable and More Accurate[J]. arXiv preprint arXiv:2002.09053, 2020.
[3] Shao S, Zhao Z, Li B, et al. Crowdhuman: A benchmark for detecting human in a crowd[J]. arXiv preprint arXiv:1805.00123, 2018.
[4] R. Quintero, I. Parra, D. F. Llorca, and M. Sotelo, "Pedestrian intention and pose prediction through dynamical models and behavior classification," in Proc of the IEEE ITSC, 2015, pp. 83–88.
[5] Pool E A I, Kooij J F P, Gavrila D M. Context-based cyclist path prediction using Recurrent Neural Networks[C]. 2019 IEEE Intelligent Vehicles Symposium (IV). IEEE, 2019: 824-830.
[6] Rasouli A, Kotseruba I, Tsotsos J K. Pedestrian Action Anticipation using Contextual Feature Fusion in Stacked RNNs[J].
[7] Rasouli A, Kotseruba I, Tsotsos J K. Are they going to cross? A benchmark dataset and baseline for pedestrian crosswalk behavior[C]. Proceedings of the IEEE International Conference on Computer Vision. 2017: 206-213.
[8] Schneemann F, Heinemann P. Context-based detection of pedestrian crossing intention for autonomous driving in urban environments[C]. 2016 IEEE/RSJ International Conference on Intelligent Robots and Systems (IROS). IEEE, 2016: 2243-2248.
[9] Deo N, Trivedi M M. Learning and predicting on-road pedestrian behavior around vehicles[C]. 2017 IEEE 20th International Conference on Intelligent Transportation Systems (ITSC). IEEE, 2017: 1-6.
[10] Radwan N, Valada A, Burgard W. Multimodal interaction-aware motion prediction for autonomous street crossing[J]. arXiv preprint arXiv:1808.06887, 2018.
[11] Alahi A, Goel K, Ramanathan V, et al. Social lstm: Human trajectory prediction in crowded spaces[C]. Proceedings of the IEEE conference on computer vision and pattern recognition. 2016: 961-971.
[12] Kooij J F P, Flohr F, Pool E A I, et al. Context-based path prediction for targets with switching dynamics[J]. International Journal of Computer Vision, 2019, 127(3): 239-262.
[13] Kooij J F P, Schneider N, Flohr F, et al. Context-based pedestrian path prediction[C]. European Conference on Computer Vision. Springer, Cham, 2014: 618-633.
[14] Schneider N, Gavrila D M. Pedestrian path prediction with recursive bayesian filters: A comparative study[C]. German Conference on Pattern Recognition. Springer, Berlin, Heidelberg, 2013: 174-183.
[15] Pavlovic, V., Rehg, J.M., MacCormick, J.: Learning switching linear models of human motion. In: Advances in NIPS. pp. 981–987 (2000)
[16] Oquab M, Bottou L, Laptev I, et al. Learning and transferring mid-level image representations using convolutional neural networks[C]. Proceedings of the IEEE conference on computer vision and pattern recognition. 2014: 1717-1724.
[17] Deng J, Dong W, Socher R, et al. Imagenet: A large-scale hierarchical image database[C]. 2009 IEEE conference on computer vision and pattern recognition. Ieee, 2009: 248-255.
[18] Mínguez R Q, Alonso I P, Fernández-Llorca D, et al. Pedestrian path, pose, and intention prediction through gaussian process dynamical models and pedestrian activity recognition[J]. IEEE Transactions on Intelligent Transportation Systems, 2018, 20(5): 1803-1814.
[19] Wang J M, Fleet D J, Hertzmann A. Gaussian process dynamical models for human motion[J]. IEEE transactions on pattern analysis and machine intelligence, 2007, 30(2): 283-298.
[20] Li J, Wang C, Zhu H, et al. Crowdpose: Efficient crowded scenes pose estimation and a new benchmark[C]. Proceedings of the IEEE Conference on Computer Vision and Pattern Recognition. 2019: 10863-10872.
[21] Fang H S, Xie S, Tai Y W, et al. Rmpe: Regional multi-person pose estimation[C]. Proceedings of the IEEE International Conference on Computer Vision. 2017: 2334-2343.
[22] Cao Z, Simon T, Wei S E, et al. Realtime multi-person 2d pose estimation using part affinity fields[C]. Proceedings of the IEEE Conference on Computer Vision and Pattern Recognition. 2017: 7291-7299.
[23] Cao Z, Hidalgo G, Simon T, et al. OpenPose: realtime multi-person 2D pose estimation using Part Affinity Fields[J]. arXiv preprint arXiv:1812.08008, 2018.
[24] Chung J, Gulcehre C, Cho K H, et al. Empirical evaluation of gated recurrent neural networks on sequence modeling[J]. arXiv preprint arXiv:1412.3555, 2014.
[25] Kingma D P, Ba J. Adam: A method for stochastic optimization[J]. arXiv preprint arXiv:1412.6980, 2014.